\documentclass[final,5p,times,twocolumn,authoryear]{elsarticle}
\usepackage{amsmath}
\usepackage{float}
\usepackage{flushend}

\usepackage{amssymb}
\journal{Computer Vision and Image Understanding}

\begin{document}

\begin{frontmatter}


 \title{Title\tnoteref{label1}}

\title{Unsupervised Total Variation Loss for Semi-supervised Deep Learning of Semantic Segmentation}


\author{Mehran Javanmardi, Mehdi Sajjadi, Ting Liu, Tolga Tasdizen}

\address{Scientific Computing and Imaging Institute, University of Utah, Salt Lake City, UT 84112}

\begin{abstract}
We introduce a novel unsupervised loss function for learning semantic segmentation with deep convolutional neural nets (ConvNet) when densely labeled training images are not available. More specifically, the proposed loss function penalizes the l1-norm of the gradient of the label probability vector image , i.e. total variation, produced by the ConvNet. This can be seen as a regularization term that promotes piecewise smoothness of the label probability vector image produced by the ConvNet during learning. The unsupervised loss function is combined with a supervised loss in a semi-supervised setting to learn ConvNets that can achieve high semantic segmentation accuracy even when only a tiny percentage of the pixels in the training images are labeled. We demonstrate significant improvements over the purely supervised setting. Furthermore, we show that using the proposed piecewise smoothness constraint in the learning phase significantly outperforms post-processing results from a purely supervised approach with Markov Random Fields.
\end{abstract}

\begin{keyword}


Semantic segmentation, Semi-supervised learning, Sparse labeling, Convolutional neural networks
\end{keyword}

\end{frontmatter}



\section{Introduction}

Semantic segmentation can be described as assigning an object label to each pixel in an image. This assignment can be binary, {\em i.e.} foreground vs. background or multi-class, {\em i.e.} \textit{"sky"}, \textit{"building"}, \textit{"horse"} etc.  Semantic segmentation is a challenging computer vision task that combines image segmentation and object detection, and can be seen as a crucial step for image understanding. Recently, ConvNets have been producing state-of-the-art results in many computer vision tasks including semantic segmentation(\cite{farabet2013learning,chen2014semantic}). However, successfully training ConvNets which can have large numbers of free parameters with existing methods requires large amounts of labeled data. It has been claimed (\cite{zhu2012we}) that improvements have a correlation not only with the methodologies used but also the amount of training data available to the algorithm. Unfortunately, building large labeled datasets can be costly and time consuming. At a time when access to unlabeled data is very easy and cheap, devising new algorithms and methods to profit from these sources is imperative. The goal of these algorithms can be to improve current state of the art results by incorporating more unlabeled data in a \textit{"transductive learning"} setting or achieve results comparable to the state of the art with significantly smaller amounts of labeled training data in a \textit{"semi-supervised learning"} paradigm which is the goal of this work. \\

Numerous methods and algorithms have been proposed for semi-supervised learning both in general and also for specific applications (\cite{chapelle2006semi,zhu2005semi}). Self-training and co-training can be mentioned as classic methods in semi-supervised learning literature (\cite{blum1998combining}). In this paper, we consider the problem of semantic segmentation when only sparse labels are available to the learning algorithm. This can be the case where a user labels a dataset only for some pixels per image which is a more realistic burden on the user than labeling all pixels in the training images. Motivated by this problem, we demonstrate how natural image characteristics, {\em i.e.} piecewise smoothness, can be used to develop novel unsupervised loss functions for learning ConvNets and demonstrate the advantages of our approach over the purely supervised setting.

\section{Related Work}

Research has demonstrated that an accurate semantic segmentation can not be achieved by only using pixel intensities of the images and it is essential to take into consideration the contextual information(\cite{torralba2004contextual,tu2010auto}). This contextual information can be defined as the relation between objects, {\em i.e.} sky above building, or the structure of an object itself, {\em i.e.} parts of the human body. Graphical models such as Conditional Random Fields (CRF) (\cite{koltun2011efficient,kohli2009robust,he2004multiscale}) and Markov Random Fields (MRF) (\cite{tighe2013superparsing,gould2009decomposing,larlus2008combining,kumar2010efficiently}) have been widely employed for semantic segmentation. CRF models consist of a unary term on each pixel or region of the image and a pairwise term that regulates consistency between neighboring pixels or regions. In \cite{he2004multiscale} a CRF has been used in a multi-scale approach to incorporate contextual information into the task of image labeling. Authors in \cite{kohli2009robust} use unsupervised methods to segment an image into regions and use a high order CRF to combine these regions. Accuracy of such methods is usually limited by the initial segmentation. To overcome this problem, CRFs have been used on the pixel level as well to produce more fine grained segmentation. In \cite{koltun2011efficient} a fully connected CRF is defined on the image which approximates the pairwise energy terms with Gaussian kernels. An MRF is utilized in \cite{gould2009decomposing} to define an energy function on both geometry and appearance of the image. They also introduce a two stage hill climbing inference technique to minimize this energy term. In \cite{tighe2013superparsing} authors implement an MRF on top of image segments to optimize the context of neighborhood regions.\\

Deep learning has recently made significant advances in the field of computer vision. ConvNets (\cite{lecun1998gradient,krizhevsky2012imagenet}) have been recently used for semantic segmentation (\cite{farabet2013learning,long2015fully,chen2014semantic,ciresan2012deep,pinheiro2013recurrent,hariharan2014simultaneous,gupta2014learning,liu2015semantic}). In \cite{ciresan2012deep} patches are extracted around each pixel and a ConvNet is trained on the patches to detect cell membranes in electron microscopy images. In \cite{farabet2013learning} authors propose a multi-scale ConvNet to extract features for scene labeling. In \cite{chen2014semantic} fully connected CRFs are incorporated to improve semantic segmentation. Authors in \cite{long2015fully} propose a Fully Convolutional Network (FCN) which alters the fully connected layers in the network to convolutional layers. FCN takes an input image of any size and outputs a probability map of the same size. FCN has shown significant improvements in accuracy for dense predictions. In the work by \cite{liu2015semantic} a CRF is applied on top of FCN which results in fine grained improvements for semantic segmentation.\\

Cascaded classifiers have also been popular for the task of semantic segmentation (\cite{torralba2004contextual,tu2010auto,fink2003mutual,heitz2009cascaded,li2010towards,seyedhosseini2013image,seyedhosseini2015semantic}). These algorithms take advantage of multiple classifiers usually in a hierarchical or multi-scale fashion to boost the accuracy of the segmentation. In \cite{tu2010auto} authors introduce a series of classifiers that integrate both information from the original image and the contextual information from previous classifier in their auto-context algorithm. In some other works (\cite{seyedhosseini2013image,seyedhosseini2015semantic}) classifiers are employed in a hierarchical fashion to incorporate contextual information from multiple scales. One significant difference between cascaded classifiers and ConvNets is that training cascaded classifiers is computationally easier due to the fact that each classifier is learned separately in an incremental fashion where ConvNets are trained in a unified manner and as one network. However, ConvNets are observed to produce slightly more accurate results than cascaded classifiers in general. \\

We should note that the Total Variation loss has been extensively used for different applications in the literature, however to the best of our knowledge it has not been used before directly in learning a discriminative network as a structured loss on the output. In \cite{houhou2009semi} authors use Total Variation loss as a regularizing factor in a continuous min­cut/max­flow approach to iteratively and interactively segment to binary images, however we use Total Variation loss as a structured loss to learn a deep network using a training set and produce segmentations on a separate testing set without any user interaction and input. Another work (\cite{li2014high}) proposes a weighted smoothness regularizer which is constrained on both the output and image itself (weighted by image closeness characterized by hand crafted features). They address the semi-­supervised segmentation problem where densely labeled images are available and the method utilizes unlabeled data with unsupervised loss to further improve accuracy whereas we address segmentation with a sparsely labeled training set which is a much harder problem. Note that access to densely labeled data promotes methods like graphical models which results in more fine grained segmentation whereas in a sparsely labeled paradigm effectiveness of such methods decreases significantly due to the lack of dense labels for training. In addition \cite{li2014high} uses a two step algorithm to train a model where in the first step predictions are made on the unlabeled images using the model and smoothness term, then in the second step the labels for unlabeled images in the first step are used as hard labels with labeled images to update the model weights, however we propose a unified image agnostic learning framework that directly backpropagates the structured loss through the network. It in noteworthy that other works like \cite{lin2016scribblesup} also utilize a similar two step algorithm to refine their segmentation at each iteration where we learn everything in a unified manner. Another work (\cite{Bearman16}) also tackles the sparse labeling problem, however their approach relies on external algorithms to provide objectness priors.

\section{Unsupervised Loss Functions from Spatial Structure}
\label{sec:unsupgeneral}

Images have spatial characteristics that can be exploited for learning semantic segmentation. For instance, regions corresponding to distinct objects in natural images are expected to be piecewise smooth whereas vasculature in medical images is curvilinear (2D) or tubular (3D). While such knowledge has been extensively used for pre-processing input images or post-processing output images in the context of semantic segmentation, the contribution of our work is their use as unsupervised loss functions directly in learning. These loss functions are constructed using spatial operations on the output images of the classifier being learned.  Therefore, multiple output image pixels are included in their computation compared to loss functions such as Mean Squared Error (MSE) where only one pixel is used to compute the error. Consequently, error backpropagation for the proposed loss functions are different. Next, we will show how these general spatial loss functions can be minimized using backpropagation. We will then derive the solution for the specific case of the total variation loss in Section~\ref{sec:tv}.

Without loss of generality consider a binary scene labeling paradigm where the goal is to assign a label \textit{"zero"} or \textit{"one"} to each pixel in an image. We have a set of training images $I_i\in\mathcal{I}$. At the pixel level, the training set consists of two sets of pixels, $\textit{(i)}$ labeled pixels which in this paper we call \textit{"supervised"} pixels, $x_k \in \mathcal{S}$ and $\textit{(ii)}$ unlabeled pixels which we call \textit{"unsupervised"} pixels, $x_k \in \mathcal{U}$. These two sets are disjoint and their union is the set of all pixels in $\mathcal{I}$.
A supervised loss function such as mean squared error 
\begin{align}
\label{eq1}
E_{\mathcal{S}} (\mathbf{w})= \sum_{x_k \in \mathcal{S}} (f(x_k;\mathbf{w}) - L(x_k))^2, 
\end{align}
where $L(x_k)$ is the training label of pixel $x_k$, can be minimized to learn the parameters $\mathbf{w}$ of the classifier $f$. However, when the number of free parameters $\mathbf{w}$ is large and the amount of labeled data available is small in comparison, we can expect overfitting and poor performance. Of course, one solution is to increase the amount of labeled data; however, this is costly in human effort as discussed previously. Regularization by penalizing the $L_2$ norm of the weight vector is another traditional approach to this problem; however, these types of regularization constraints are very general and do not exploit the characteristics of the spatial structures present in images.\\

Let $P_i = f(\mathcal{I}_i)$ denote the probability map generated by the classifier for the training image $\mathcal{I}_i$ and let $\mathcal{P}$ denote the set of all such probability images for our training data. Let 
\begin{align}
\label{eq2}
E_{\mathcal{U}} (\mathbf{w}) = \sum_{x_k \in \mathcal{I}} \Theta\left(\overline{f}(x_k)\right)
\end{align}
where $\overline{f}(x_k)$ is a vector of size $n^2$ containing the $n \times n$ neighborhood of pixel $x_k$ in the probability map image $P_i$ and $\Theta:\mathbb{R}^{n^2} \rightarrow \mathbb{R}$ is a function which computes some spatial characteristic of the probability map image which we want to penalize. 
We should point out that labeled data can be treated as unlabeled data too, therefore we consider the whole set $\mathcal{I}$ for defining the unsupervised loss. Consider an offset vector $\gamma$ which defines the spatial offset of the $n \times n$ neighborhood with respect to the $k$th pixel $x_k$. In this case we will have,
\begin{align}
\label{eq3}
\overline{f}(x_k) = \begin{bmatrix}
f_1(x_k)\\ 
f_2(x_k)\\ 
\dots \\ 
f_{n^2}(x_k)
\end{bmatrix}_{n^2 \times 1}
\end{align}
where we have,
\begin{align}
\label{eq3.1}
f_i(x_k) = f(x_{k+\gamma(i)}).
\end{align}
We note that $E_{\mathcal{U}}$ is a function of $f_i(x_k)$ and $i = 1, 2, \dots, n^2$ which are themselves functions of \textbf{w}
\begin{align}
\label{eq4}
E_{\mathcal{U}}(\textbf{w}) = \sum_{x_k \in \mathcal{I}} \Theta\left( f_1(x_k; \textbf{w}), f_2(x_k; \textbf{w}), \dots, f_{n^2}(x_k; \textbf{w})  \right).
\end{align}
We adopt stochastic gradient descent  to minimize this loss function(\cite{bottou1991stochastic,lecun2012efficient}). By applying the multivariate chain rule to compute the gradient of $E_{\mathcal{U}}$ with respect to \textbf{w} we have,
\begin{align}
\label{eq5}
\frac{\partial E_{\mathcal{U}}}{\partial \textbf{w}} = \sum_{x_k \in \mathcal{I}} \sum_{i = 1}^{n^2} \frac{\partial E_{\mathcal{U}}}{\partial f_i(x_k)} \times \frac{d f_i(x_k)}{d \textbf{w}}
\end{align}
where $\frac{d f_i(x_k)}{d \textbf{w}}$ is the derivative of $f(x)$ with respect to $\mathbf{w}$ computed at $x = x_{k+\gamma(i)}$ and $\frac{\partial E_{\mathcal{U}}}{\partial f_i(x_k)}$ is a coefficient which is defined by the specific choice of the unsupervised loss function. In other words, the backpropagation of a spatial loss function is a linear combination of the derivatives of $f$ with respect to the weights calculated at specific locations with respect to a specific sample pixel. The coefficients of this linear combination are dependent on the definition of the loss function. Finally, note that the proposed unsupervised loss can be combined with a supervised loss $E_{\mathcal{S}} + \alpha E_{\mathcal{U}}$ to work in a semi-supervised learning setting, where $\alpha$ is a hyper parameter that controls the effect of the unsupervised term and is obtained by cross validation.

\section{Total Variation Loss for Learning}
\label{sec:tv}

The piecewise smoothness property of natural images has been used extensively in the fields of image processing and computer vision. In \cite{perona1990scale} a nonlinear PDE is introduced to remove noise from the image while conserving most details of the original image with the assumption that image intensities are piecewise flat. This diffusion results in piecewise homogeneous regions in the image which is achieved by controlling the diffusion around edges. In a similar work (\cite{rudin1992nonlinear}) it is proposed to minimize the total variation of an image to suppress the noise and preserve the edges at the same time. We observe that natural scene image segmentation probability maps should result in piecewise smooth regions which means that if a pixel has a high probability of belonging to a specific object category, its neighboring pixels also have high probability to belong to the same category. This suggests that a probability map that generates an optimum segmentation of a natural scene image should have low total variation. This observation motivates us to use total variation of probability map images as an unsupervised penalty to help improve segmentation results when only sparsely labeled images are available. We define the unsupervised loss function as

\begin{eqnarray}
\label{eq6}
E_{\mathcal{U}}(\textbf{w}) &=& \sum_{x_k \in \mathcal{I}} \Theta\left(\overline{f}(x_k; \textbf{w})\right) = \sum_{x_k \in \mathcal{I}} |\nabla f(x_k; \textbf{w}) |\\
&=& \sum_{x_k \in \mathcal{I}} |\nabla_X f(x_k)| + | \nabla_Y f(x_k)|.
\label{eq8}
\end{eqnarray}
Although equation (\ref{eq8}) is convex, it is not differentiable. We use the subgradient method which generalizes the derivative to non-differentiable functions to compute
\begin{eqnarray}
\frac{\partial E_{\mathcal{U}}}{\partial \textbf{w}} &=& \sum_{x_k \in \mathcal{I}} \frac{d}{d \textbf{w}} |\nabla_X f(x_k)| + \frac{d}{d \textbf{w}} |\nabla_Y f(x_k)| 
\label{eq9}
\\
&=& \sum_{x_k \in \mathcal{I}} \text{sign}(\nabla_X f(x_k)) \frac{d}{d \textbf{w}} \nabla_X f(x_k)\\
\nonumber &+&\text{sign}(\nabla_Y f(x_k))\frac{d}{d \textbf{w}} \nabla_Y f(x_k).
\label{eq10}
\end{eqnarray}
We use the Sobel operators
\begin{align}
\label{eq7}
\nabla_X = \begin{bmatrix}
-1 & -2 & -1 \\ 
0 & 0 & 0\\ 
1 & 2 & 1
\end{bmatrix}, \, \, \, \nabla_Y = \begin{bmatrix}
-1 & 0 & 1\\ 
-2 & 0 & 2\\ 
-1 & 0 & 1
\end{bmatrix}
\end{align}
to calculate the derivatives of the probability map image in $X$ and $Y$ directions, respectively. 
These derivative computations can be written as the vector products 
\begin{align}
\label{eq11}
\nabla_X f(x_k) = \overline{\nabla_X}^T \overline{f}(x_k), \nabla_Y f(x_k) = \overline{\nabla_Y}^T  \overline{f}(x_k),
\end{align}
where $\overline{\nabla_X}$ and $\overline{\nabla_Y}$ are the vectorized versions of $\nabla_X$ and $\nabla_Y$ operators in (\ref{eq7}) in row major form, respectively, and the superscript $T$ denotes vector transpose. Let $\gamma(i)$, $i = 1, \dots, 9$, be the offsets in (\ref{eq3.1}) chosen to represent the pixels in the $3\times 3$ neighborhood in a row major fashion beginning from top left and ending in bottom right. We then have
\begin{align}
\label{eq12}
\nabla_X f(x_k) =\begin{bmatrix}
-1\\ 
-2\\ 
-1\\ 
0\\ 
0\\ 
0\\ 
1\\ 
2\\ 
1
\end{bmatrix}^T\begin{bmatrix}
f_1(x_k)\\ 
f_2(x_k)\\ 
f_3(x_k)\\ 
f_4(x_k)\\ 
f_5(x_k)\\ 
f_6(x_k)\\ 
f_7(x_k)\\ 
f_8(x_k)\\ 
f_9(x_k)
\end{bmatrix}, \nabla_Y f(x_k) = \begin{bmatrix}
-1\\ 
0\\ 
1\\ 
-2\\ 
0\\ 
2\\ 
-1\\ 
0\\ 
1
\end{bmatrix}^T\begin{bmatrix}
f_1(x_k)\\ 
f_2(x_k)\\ 
f_3(x_k)\\ 
f_4(x_k)\\ 
f_5(x_k)\\ 
f_6(x_k)\\ 
f_7(x_k)\\ 
f_8(x_k)\\ 
f_9(x_k)
\end{bmatrix}
\end{align}
where $f_i(x_k)$ is defined as in equation (\ref{eq3.1}). Both spatial derivative operators $\nabla_X$ and $\nabla_Y$ and also the derivative operator $\frac{d}{d \textbf{w}}$ are linear and interchangeable.  Interchanging the order of differentiation with respect to $\mathbf{w}$ and the differentiation with respect to $X$ and $Y$ in (\ref{eq10}), and substituting the expressions for the spatial derivatives into (\ref{eq10}), we obtain
\begin{align}
\label{eq13}
\frac{\partial E_{\mathcal{U}}}{\partial \textbf{w}} &= \sum_{x_k \in \mathcal{I}} \sum_{i = 1}^9 \text{sign}(\nabla_X f(x_k))  
\overline{\nabla_X}(i) \frac{d f_i(x_k)}{d\textbf{w}}\\ \nonumber
& \quad \quad + \text{sign}(\nabla_Y f(x_k))  \overline{\nabla_Y}(i) \frac{d f_i(x_k)}{d\textbf{w}}
\end{align}
where $\overline{\nabla_X}(i)$ and $\overline{\nabla_Y}(i)$ are the $i$th element in vectors $\overline{\nabla_X}$ and $\overline{\nabla_Y}$, respectively. In the last section we derived the gradient for a general spatial penalty on the probability map image and obtained equation (\ref{eq5}). Comparing equation (\ref{eq5}) with equation (\ref{eq13}) which shows the gradient of the total variation penalty we can observe that
\begin{align}
\label{eq14}
\frac{\partial E_{\mathcal{U}}}{\partial f_i(x_k)} = \text{sign}(\nabla_X f(x_k)) \times \overline{\nabla_X}(i) + \text{sign}(\nabla_Y f(x_k)) \times \overline{\nabla_Y}(i)
\end{align}
The gradient of the unsupervised penalty at $x_k$ is defined by a linear combination of gradient values of the system at $x_k$'s eight connected neighborhood pixels. This implies a dependency between pixels in the probability map image which is the goal of total variation penalty.\\

\section{Experiments}
\label{sec:experiments}
In this section, we use two different pixel classifiers, Contextual Hierarchical Model (CHM) (\cite{seyedhosseini2015semantic}) and ConvNet pixel classifier (\cite{ciresan2012deep}), to show that the proposed loss function can improve accuracy regardless of the specific classifier choice. CHM (\cite{seyedhosseini2015semantic}) is a cascaded pixel classifier which combines multiple classifiers in a hierarchical fashion to boost its accuracy. CHM consists of two stages, bottom-up and top-down. In the bottom-up stage CHM creates a multi-scale pyramid of contextual maps which are generated by applying a classifier at each scale on the downsampled original image and the previous contextual maps in the pyramid. As we go to higher levels in the bottom-up stage, the resolution of the image decreases which means the higher level classifier is able to capture more global features than a lower level classifier which captures more fine-grained details. Finally, all the contextual maps in the pyramid created at the bottom-up stage and the image features from the original image are used in the top-down stage classifier to compute the final probability map image. We use CHM as our learning system for experiments on the Weizmann horse dataset.\\

We also employ the ConvNet proposed in (\cite{ciresan2012deep}) for pixel classification, with some modifications, to show practicality of our unsupervised penalty on ConvNets. In this model we extract a patch around each pixel of the image and train a ConvNet to classify that pixel. In the original work by Ciresan et al, multiple data augmentations and averaging of different models are used to get state of the art results. Since the goal of this paper is not achieving state of the art results, but to show improvements in a semi-supervised paradigm we have not performed any data augmentation or classifier averaging. We used a ConvNet consisting of 4 convolutional layers containing 64 maps and a kernel size of $3 \times 3$, each one followed by a ReLU layer, a maxpooling layer and a Local Response Normalization (LRN) layer and is continued with a fully connected layer of size 512. Finally, a softmax layer is used to generate probabilities for each class in the dataset. We used Berkeley Vision and Learning Center's Caffe (\cite{jia2014caffe}) framework to implement this ConvNet.

\subsection{Weizmann Horse Data Set}

The Weizmann Horse data set (\cite{borenstein2008combined}) consists of 328 gray scale images of horses with different sizes provided with pixel labels for background and foreground. We followed the approach by (\cite{tu2010auto,seyedhosseini2015semantic}) and use half of the images as training and the other half for testing. Note that we are considering a setting where only sparse labels are available. We randomly choose a specific number of labeled pixels from each training image and consider the rest of the pixels as unlabeled training data to create sparse label sets for training. We randomly created sparse label training sets of sizes 10, 20, 30, 40, and 50 labeled pixels per training image. We generate five different sparse label training sets for each size and report the average and standard deviation on the testing data. \\

For each of these sets we trained two different CHMs with 3 levels in the bottom-up stage. One CHM was trained with the supervised penalty and the other one with both the supervised penalty and unsupervised penalty. Note that we only applied the unsupervised penalty to the classifiers in CHM that work on the original resolution of the image and not the classifiers that have lower resolution. We applied the penalty to both the first classifier of the bottom-up stage and the classifier of the top-down stage. The reasoning behind this is the fact that CHM tends to generate smooth probability maps in coarser resolutions inherently and applying the total variation to those levels does not improve the results significantly. 

\setlength{\tabcolsep}{4pt}
\begin{table*}[t]
\begin{center}
\caption{Pixel error results for Weizmann Horse dataset. The reported numbers are average of 5 experiments and corresponding standard deviation. Average(\%)$\pm$ std dev.}
\label{wz}
\begin{tabular}{cccc}
\hline\noalign{\smallskip}
\# of labeled pixels per image  & supervised approach & MRF post processing & semi-supervised approach\\
(\% of whole training set)\\
\noalign{\smallskip}
\hline
\noalign{\smallskip}
10 (0.0168 \%)& $23.81 \pm 1.50 $ & $ 19.92 \pm 1.28 $ & $ 11.98 \pm 0.24 $ \\
20 (0.0336 \%)& $16.63 \pm 0.67 $ & $ 14.24 \pm 0.36 $ & $ 11.03 \pm 0.21 $ \\
30 (0.0504 \%)& $14.20 \pm 0.80 $ & $ 12.27 \pm 0.29 $ & $ 10.30 \pm  0.41 $ \\
40 (0.0671 \%)& $12.35 \pm 0.16 $ & $ 11.21 \pm 0.12 $ & $ 9.96 \pm 0.35 $ \\
50 (0.0839 \%)& $11.86 \pm 0.19 $ & $ 11.01 \pm 0.17 $ & $ 9.78 \pm 0.20 $\\
\hline
\end{tabular}
\end{center}
\end{table*}
\setlength{\tabcolsep}{1.4pt}

\begin{figure}[h!]
\centering
\includegraphics[height=7cm]{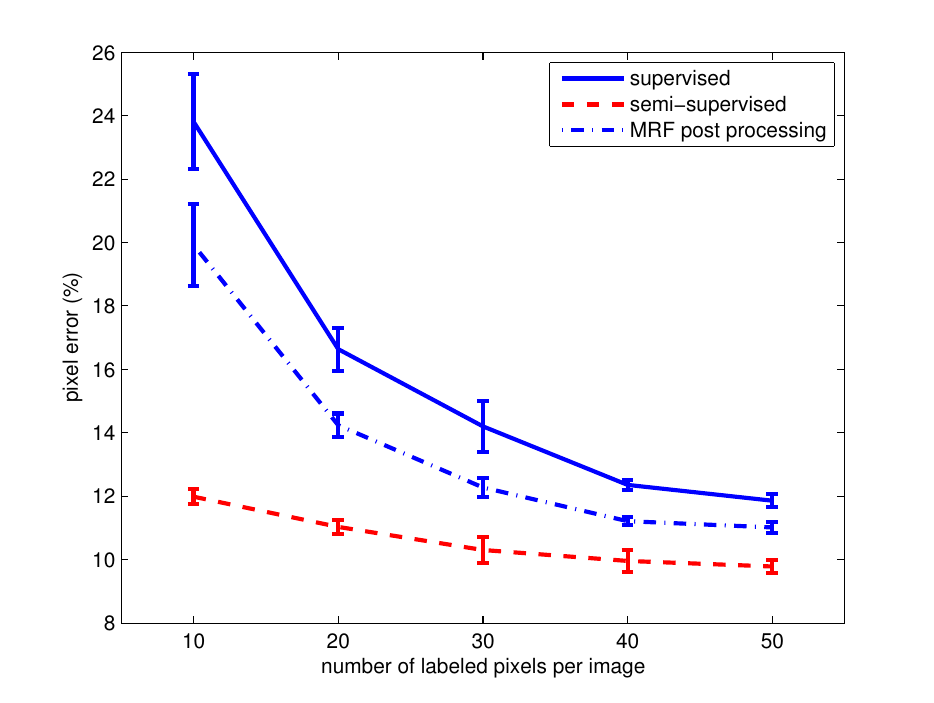}
\caption{Mean testing error plot for the Weizmann Horse dataset. Bars indicate standard deviation over the five trials at each size.}
\label{wz1}
\end{figure}

\begin{figure*}[h!]
\centering
\begin{tabular}{ccccc}
{\includegraphics[width = 1.3in]{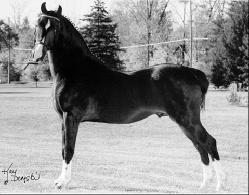}} &
{\includegraphics[width = 1.3in]{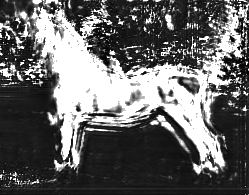}} &
{\includegraphics[width = 1.3in]{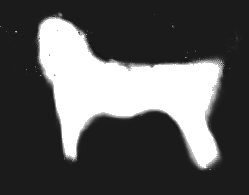}} &
{\includegraphics[width = 1.3in]{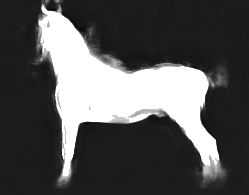}}&
{\includegraphics[width = 1.3in]{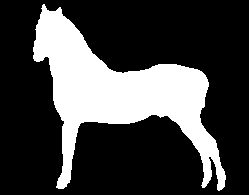}}\\
{\includegraphics[width = 1.3in]{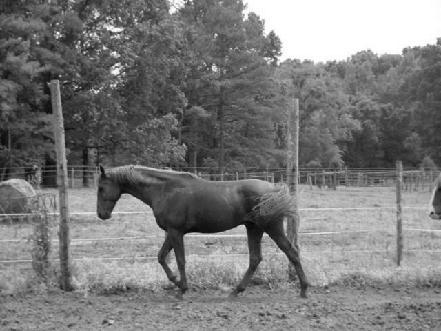}} &
{\includegraphics[width = 1.3in]{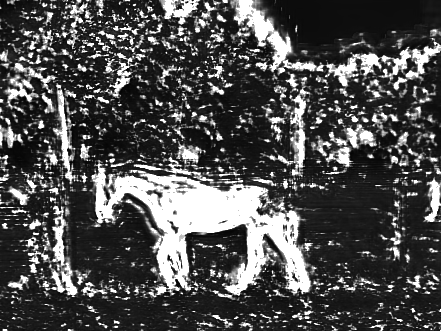}} &
{\includegraphics[width = 1.3in]{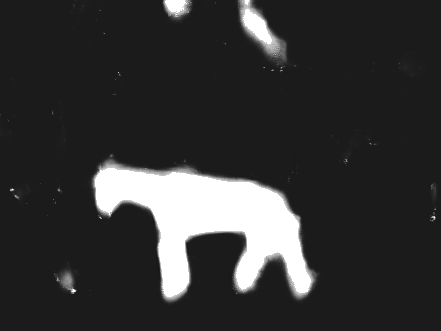}} &
{\includegraphics[width = 1.3in]{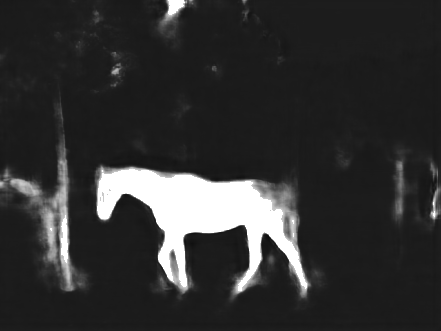}}&
{\includegraphics[width = 1.3in]{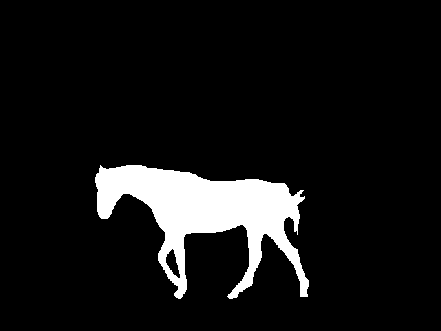}}\\
{\includegraphics[width = 1.3in]{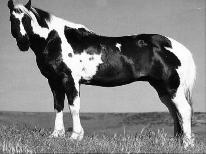}} &
{\includegraphics[width = 1.3in]{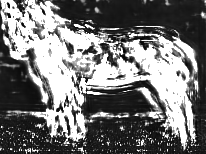}} &
{\includegraphics[width = 1.3in]{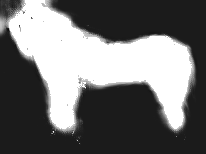}} &
{\includegraphics[width = 1.3in]{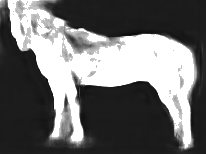}}&
{\includegraphics[width = 1.3in]{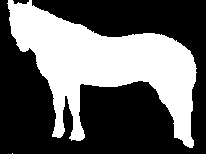}}\\
{\includegraphics[width = 1.3in]{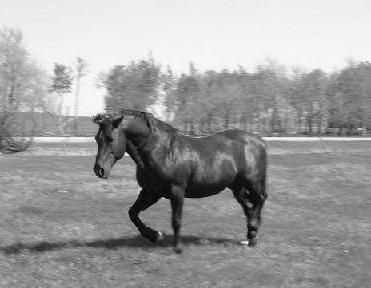}} &
{\includegraphics[width = 1.3in]{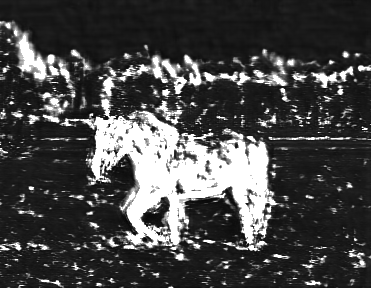}} &
{\includegraphics[width = 1.3in]{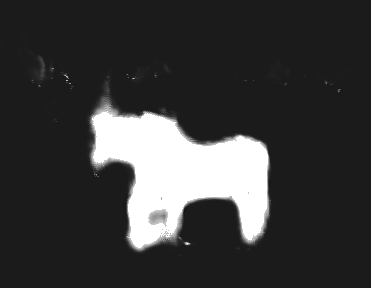}} &
{\includegraphics[width = 1.3in]{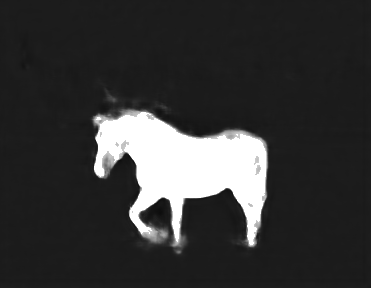}}&
{\includegraphics[width = 1.3in]{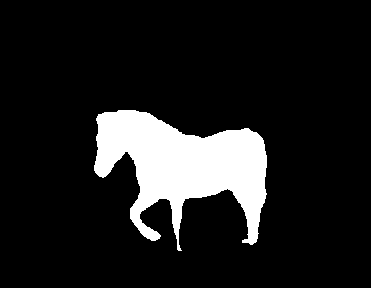}}
\end{tabular}
\caption{Sample images from Weizmann Horse dataset, using only 10 labeled pixel per image for training. The first column shows the original image, the second column are the output of the purely supervised method and the third column images are the results of the proposed method. The fourth column shows the probability maps of the model using dense (full) labels for training and the last column shows the ground truth images.}
\label{wz2}
\end{figure*}

The results on the testing dataset can be seen in Table (\ref{wz}), Figures (\ref{wz1}) and (\ref{wz2}). The total variation loss we proposed for learning improves the results significantly, especially when we have very sparse labeled data available. It is noteworthy that the CHM used for this experiment achieves a pixel error of $5.73 \%$ when fully labeled data is used to train the model. The state of the art pixel error on this dataset is $4.1 \%$ (\cite{liu2007hybrid}). Therefore, somewhat better results can be expected if our proposed unsupervised loss function is applied to the model in (\cite{liu2007hybrid}). We also note that using the proposed total variation loss does not significantly effect the results when the full set of pixels in the training set is treated as the supervised set. This is expected as the characteristics of spatial structures in the training images can be learned directly from the fully labeled data in this case.\\

Since the proposed unsupervised loss function can be seen as a regularization term in learning, we compared to classical Tikhonov regularization by penalizing the $L_2$ norm of the weight vector $\mathbf{w}$. However, the results obtained with this approach were not significantly better than the purely supervised learning without regularization. \\

Finally, we also demonstrate the advantages of imposing spatial constraints in the learning phase vs. the inference phase. Therefore, we used the UGM framework (\cite{schmidt2012ugm}) to implement an MRF that was used to post-process the resulting probability images from the supervised only approach. Although the accuracy was increased by MRF post-processing (see Table~\ref{wz1}), the proposed total variation loss has much lower error rates. \\

\subsection{Stanford Background Data Set}

The Stanford Background dataset (\cite{gould2009decomposing}) contains 715 RGB images and the corresponding label images. Images are approximately $240 \times 320$ pixels in size and pixels are classified into eight different categories. We follow the standard evaluation procedure for this dataset and do a five-fold cross validation. We randomly divide the images into 5 separate bins, each containing 143 images. We train on 4 of these bins and test on the other one and repeat the procedure for all 5 bins. \\

\setlength{\tabcolsep}{4pt}
\begin{table*}
\begin{center}
\caption{Pixel error results for Stanford Background dataset. The reported numbers are average of 5 experiments and corresponding standard deviation. Average(\%)$\pm$ std dev. }
\label{sbd}
\begin{tabular}{cccc}
\hline\noalign{\smallskip}
\# of labeled pixels per image  & supervised approach & MRF post processing & semi-supervised approach\\
(\% of whole training set)\\
\noalign{\smallskip}
\hline
\noalign{\smallskip}
10 (0.0130 \%) & $ 39.98 \pm 1.35 $ & $ 40.92 \pm 1.73 $ & $ 26.19 \pm 0.33 $\\
20 (0.0260 \%) & $ 34.35 \pm 0.98 $ & $ 34.64 \pm 1.03 $ & $ 24.33 \pm 0.29 $\\
30 (0.0391 \%) & $ 31.36 \pm 0.85 $ & $ 31.58 \pm 0.83 $ & $ 23.61 \pm  0.25 $\\
40 (0.0521 \%) & $ 29.76 \pm 0.38 $ & $ 29.93 \pm 0.36 $ & $ 23.13\pm 0.16 $\\
50 (0.0651 \%) & $ 27.96 \pm 0.60 $ & $ 28.13 \pm 0.62 $ & $ 22.79 \pm 0.24 $\\
\hline
\end{tabular}
\end{center}
\end{table*}
\setlength{\tabcolsep}{1.4pt}

\begin{figure}[h!]
\centering
\includegraphics[height=7cm]{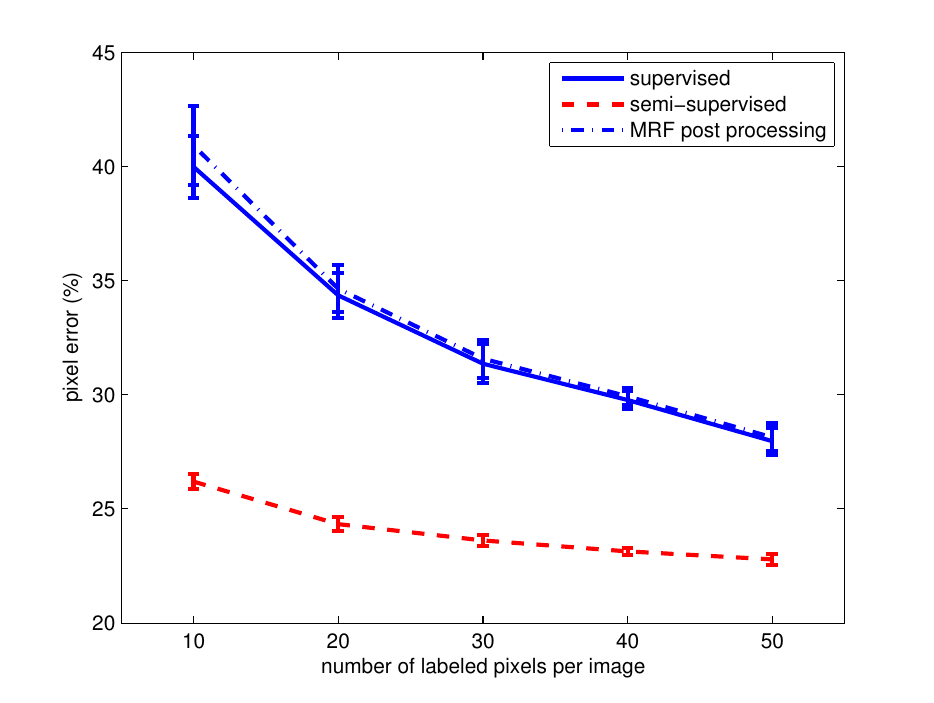}
\caption{Mean testing error plot for the Stanford background dataset. Bars indicate standard deviation over the five trials at each size.}
\label{sbd1}
\end{figure}

% Note: capitalize Figure (instead of figure) same with Table
We randomly created sparse label training sets of sizes 10, 20, 30, 40, and 50 labeled pixels per training image. We generate five different sparse label training sets for each size and report the average and standard deviation on the testing data. We use the ConvNet explained in the beginning of Section~\ref{sec:experiments} as the pixel classifier in this experiment. The mean testing pixel errors and standard deviations can be seen in Table (\ref{sbd}) and Figure (\ref{sbd1}). We post processed results of purely supervised leaning with a simple MRF, and as can be observed from Table (\ref{sbd}) and Figure (\ref{sbd1}) there is not significant improvement in the results. The supervised learning results are too noisy and even in some cases incorrect that an MRF cannot make significant corrections. Note that training better MRF and CRF models also is contingent on availability of dense labeled training images. An important advantage of our method is its ability to work in a sparsely labeled training image paradigm. 

\begin{figure*}[h!]
\centering
\begin{tabular}{ccccc}
{\includegraphics[width = 1.3in]{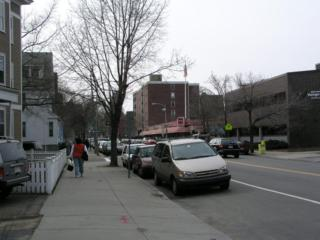}} &
{\includegraphics[width = 1.3in]{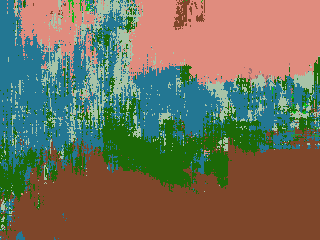}} &
{\includegraphics[width = 1.3in]{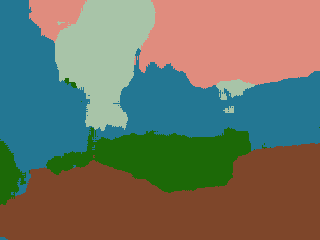}} &
{\includegraphics[width = 1.3in]{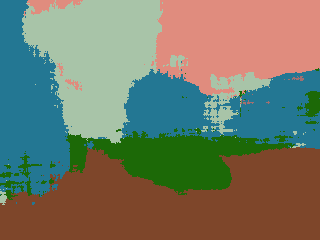}} &
{\includegraphics[width = 1.3in]{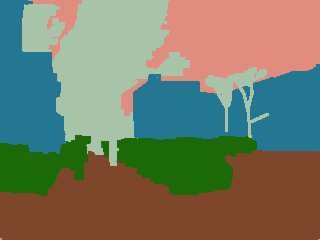}}\\
{\includegraphics[width = 1.3in]{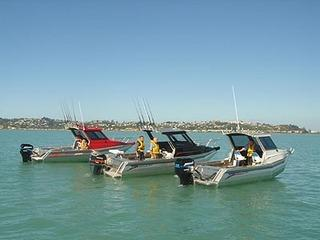}} &
{\includegraphics[width = 1.3in]{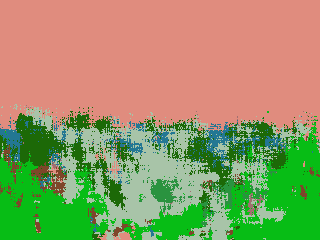}} &
{\includegraphics[width = 1.3in]{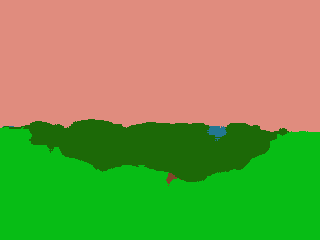}} &
{\includegraphics[width = 1.3in]{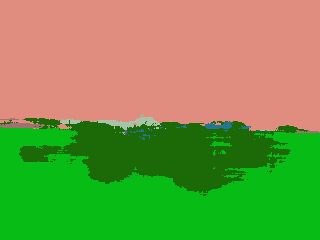}} &
{\includegraphics[width = 1.3in]{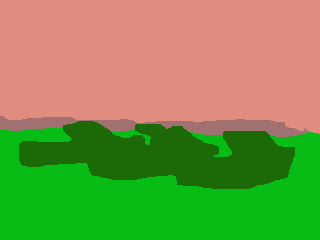}}\\
{\includegraphics[width = 1.3in]{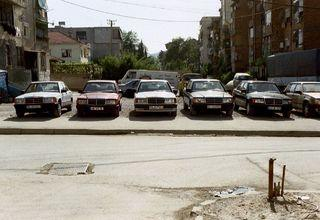}} &
{\includegraphics[width = 1.3in]{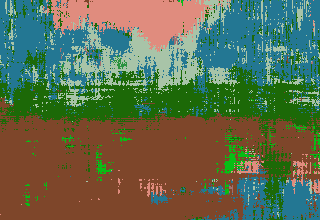}} &
{\includegraphics[width = 1.3in]{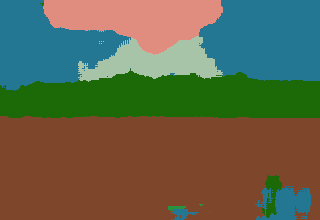}} &
{\includegraphics[width = 1.3in]{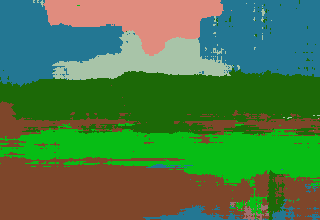}} &
{\includegraphics[width = 1.3in]{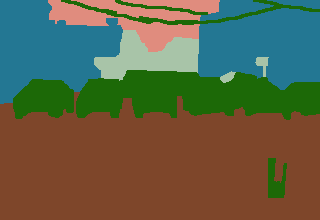}}\\
{\includegraphics[width = 1.3in]{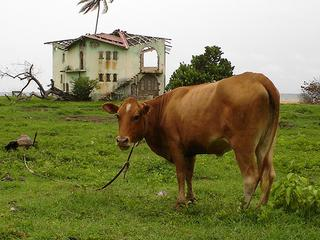}} &
{\includegraphics[width = 1.3in]{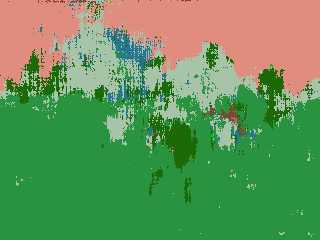}} &
{\includegraphics[width = 1.3in]{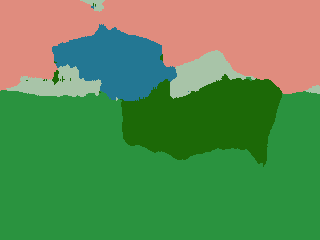}} &
{\includegraphics[width = 1.3in]{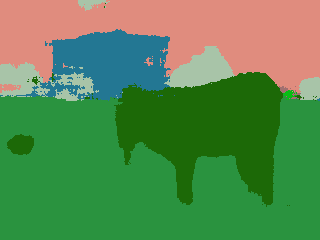}} &
{\includegraphics[width = 1.3in]{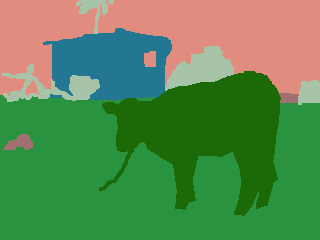}}\\
{\includegraphics[width = 1.3in]{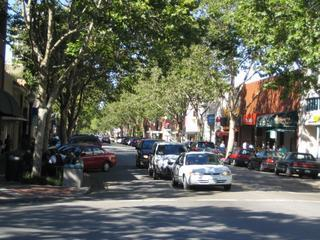}} &
{\includegraphics[width = 1.3in]{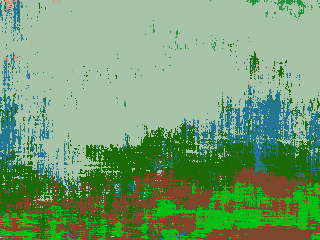}} &
{\includegraphics[width = 1.3in]{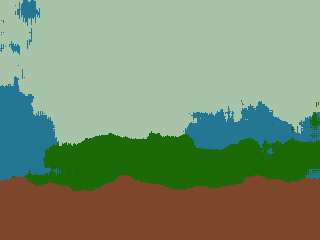}} &
{\includegraphics[width = 1.3in]{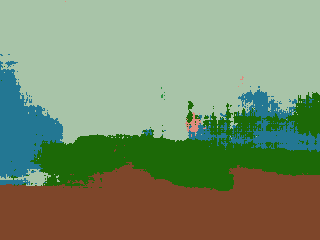}} &
{\includegraphics[width = 1.3in]{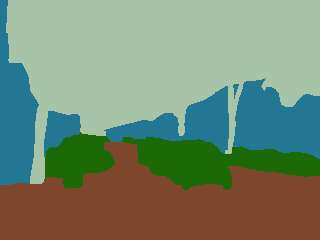}}
\end{tabular}
\caption{Sample images from Stanford Background dataset, using only 10 labeled pixel per image for training. The first column shows the original image, the second column are the output of the purely supervised method and the third column images are the results of the proposed method. The fourth column shows the probability maps of the model using dense (full) labels for training and the last column shows the ground truth images.}
\label{sbd2}
\end{figure*}

Table (\ref{sbd_2}) reports the average per-class accuracy on Stanford Background dataset. As can be seen the proposed approach consistently improves the average per-class accuracy on a multi-class dataset.

\setlength{\tabcolsep}{4pt}
\begin{table}[H]
\begin{center}
\caption{Average per-class accuracy for Stanford Background dataset. The reported numbers are average of 5 experiments }
\label{sbd_2}
\begin{tabular}{ccc}
\hline\noalign{\smallskip}
\# of labeled pixels  & supervised & semi-supervised \\
 per image & approach & approach \\
\noalign{\smallskip}
\hline
\noalign{\smallskip}
10  & $ 51.73  $ & $ 63.95  $ \\
20  & $ 56.97  $ & $ 65.91  $ \\
30  & $ 60.23  $ & $ 67.19  $ \\
40  & $ 61.70  $ & $ 67.03 $ \\
50  & $ 63.68  $ & $ 67.38  $ \\
\hline
\end{tabular}
\end{center}
\end{table}
\setlength{\tabcolsep}{1.4pt}

For comparison, we also trained a ConvNet where we use full labels of the dataset. We achieved a pixel error of $20.55 \%$ after one epoch over all the pixels in the training set, where the state of the art pixel error for this data set (\cite{seyedhosseini2015semantic}) is $17.05 \%$. Visual examples are given in Figure (\ref{sbd2}). Note that while the purely supervised learning results in a classifier whose output label maps are noisy whereas the proposed approach results in a classifier which generates piecewise smooth labeled regions and is closer visually to supervised learning with the dense label set.

\subsection{Sift Flow Data Set}

\begin{figure}[h!]
\centering
\includegraphics[height=7cm]{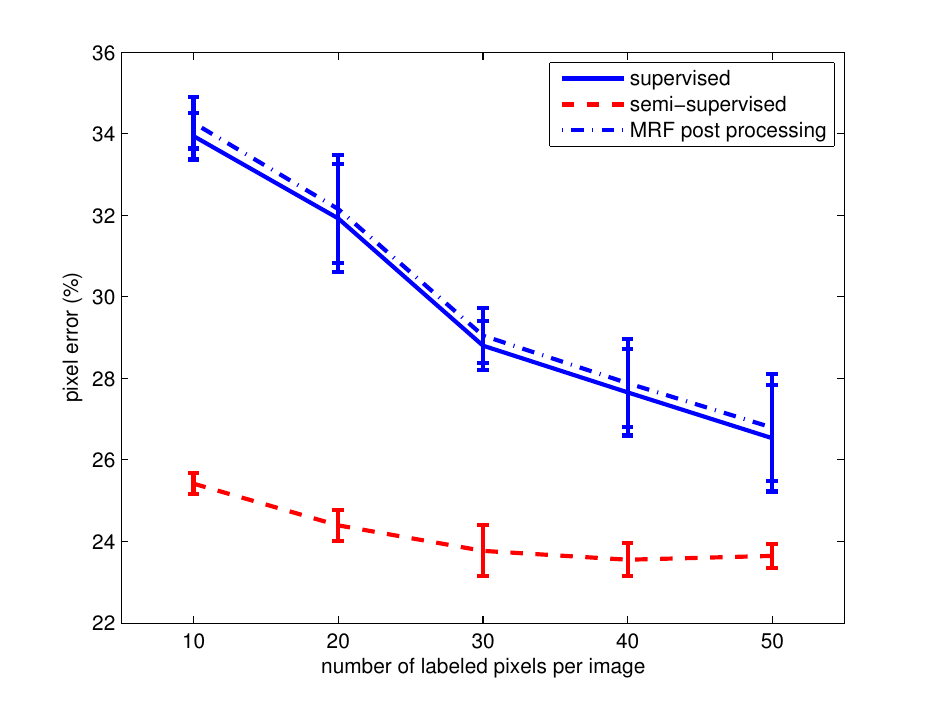}
\caption{Mean testing error plot for the Sift Flow dataset. Bars indicate standard deviation over the five trials at each size.}
\label{sf1}
\end{figure}

The Sift Flow dataset (\cite{liu2011nonparametric}) contains 2488 training images and 200 testing images all of size $256 \times 256$. It contains 33 different classes of which only 30 of them are present in the test set. We use the same ConvNet used for evaluating the Stanford Background dataset as our pixel classifier. We follow the same procedure as before to generate sparse label sets. The results for the Sift Flow data set are given in Table (\ref{sf}) and Figure (\ref{sf1}). We also post processed supervised results with an MRF to increase consistency but it did not make any significant changes in the results. Note that the MRFs we used on these datasets to improve label consistency are not trained with any dense labels since we are considering a setting where only sparse labels are available. Therefore, it is expected that MRFs will not result in significant improvement, especially in the multi-class cases where dense label information can be more informative to the MRF.\\

We also trained our model using dense labels where we achieved a pixel error of $20.78 \%$. The state of the art pixel error on this data set before the significant improvement of Long {\em et al.} (\cite{long2015fully}) to a $14.8 \%$ pixel error was achieved by Tighe {\em et al.} (\cite{tighe2013finding}) with a pixel error of $21.4 \%$.

\setlength{\tabcolsep}{4pt}
\begin{table*}
\begin{center}
\caption{Pixel error results for Sift Flow dataset. The reported numbers are average of 5 experiments and corresponding standard deviation. Average(\%)$\pm$ std dev. }
\label{sf}
\label{table:headings}
\begin{tabular}{cccc}
\hline\noalign{\smallskip}
\# of labeled pixels per image  & supervised approach & MRF post processing & semi-supervised approach\\
(\% of whole training set)\\
\noalign{\smallskip}
\hline
\noalign{\smallskip}
10 (0.0153 \%) & $33.94 \pm 0.57 $ & $ 34.26 \pm 0.62 $ & $ 25.41 \pm 0.25 $\\
20 (0.0305 \%) & $31.92 \pm 1.32 $ & $ 32.15 \pm 1.31 $  & $ 24.39 \pm 0.38 $\\
30 (0.0458 \%) & $28.80 \pm 0.60 $ & $ 29.05 \pm 0.68 $  & $ 23.77 \pm  0.62 $\\
40 (0.0610 \%) & $27.66 \pm 1.05 $ & $ 27.88 \pm 1.08 $  & $ 23.55 \pm 0.40 $\\
50 (0.0763 \%) & $26.53 \pm 1.30 $ & $ 26.79 \pm 1.31 $  & $ 23.64 \pm 0.29 $\\
\hline
\end{tabular}
\end{center}
\end{table*}
\setlength{\tabcolsep}{1.4pt}

\begin{figure*}[h!]
\centering
\begin{tabular}{ccccc}
{\includegraphics[width = 1.3in]{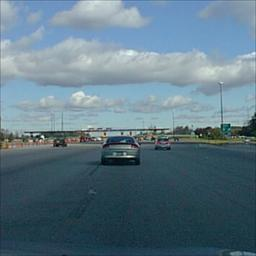}} &
{\includegraphics[width = 1.3in]{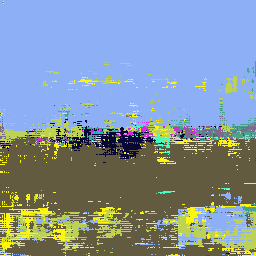}} &
{\includegraphics[width = 1.3in]{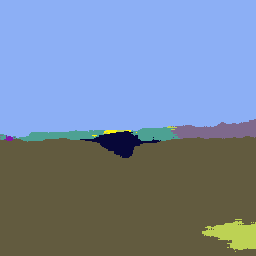}} &
{\includegraphics[width = 1.3in]{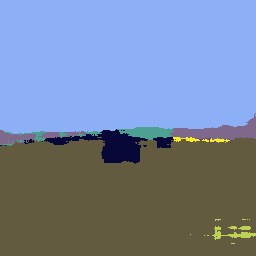}} &
{\includegraphics[width = 1.3in]{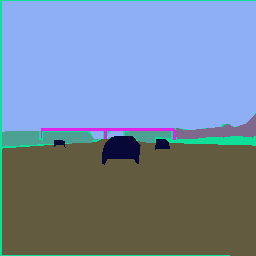}}\\
{\includegraphics[width = 1.3in]{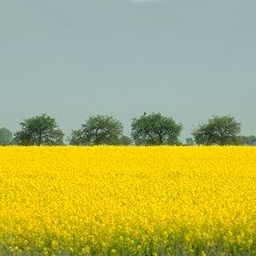}} &
{\includegraphics[width = 1.3in]{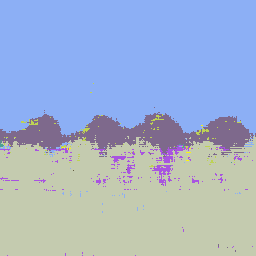}} &
{\includegraphics[width = 1.3in]{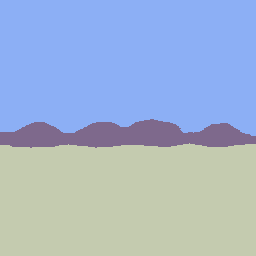}} &
{\includegraphics[width = 1.3in]{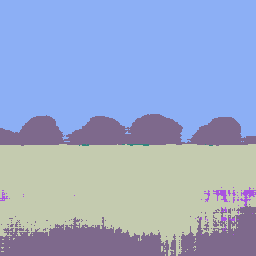}} &
{\includegraphics[width = 1.3in]{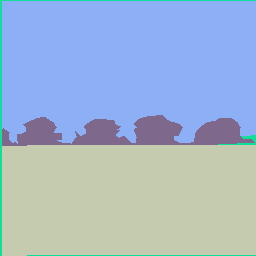}}\\
{\includegraphics[width = 1.3in]{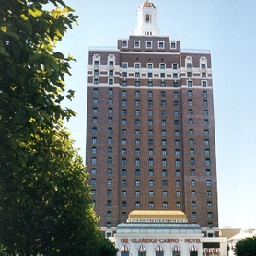}} &
{\includegraphics[width = 1.3in]{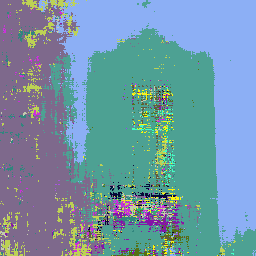}} &
{\includegraphics[width = 1.3in]{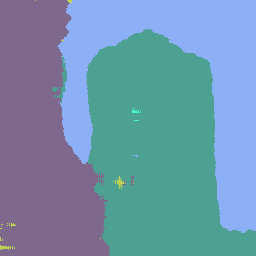}} &
{\includegraphics[width = 1.3in]{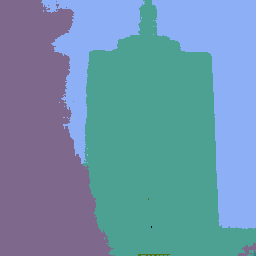}} &
{\includegraphics[width = 1.3in]{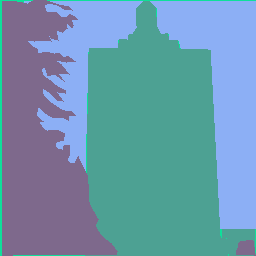}}\\
{\includegraphics[width = 1.3in]{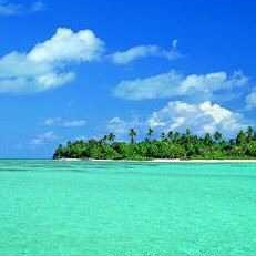}} &
{\includegraphics[width = 1.3in]{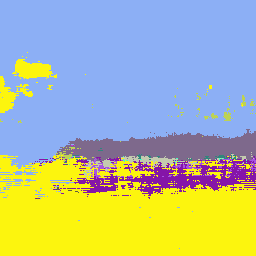}} &
{\includegraphics[width = 1.3in]{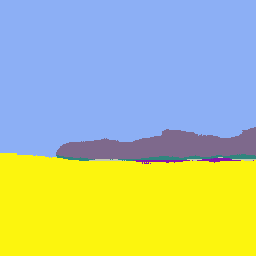}} &
{\includegraphics[width = 1.3in]{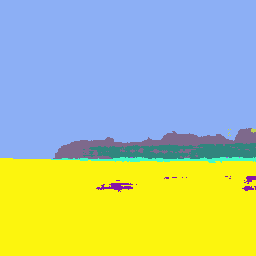}} &
{\includegraphics[width = 1.3in]{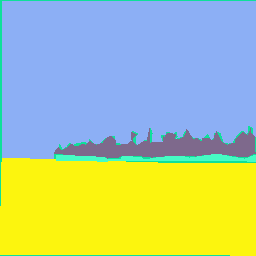}}\\
{\includegraphics[width = 1.3in]{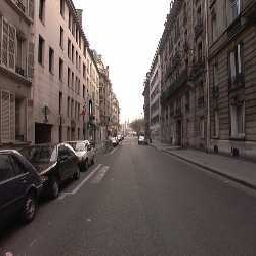}} &
{\includegraphics[width = 1.3in]{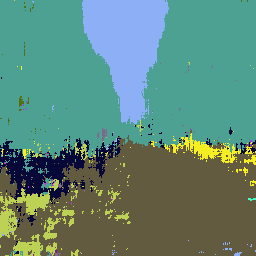}} &
{\includegraphics[width = 1.3in]{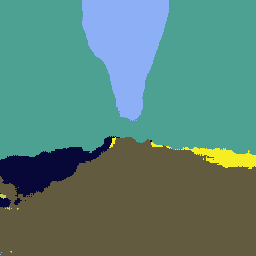}} &
{\includegraphics[width = 1.3in]{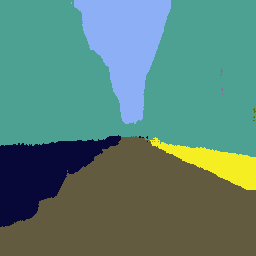}} &
{\includegraphics[width = 1.3in]{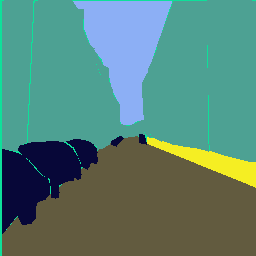}}
\end{tabular}
\caption{Sample images from Sift Flow dataset, using only 10 labeled pixel per image for training. The first column shows the original image, the second column are the output of the purely supervised method and the third column images are the results of the proposed method. The fourth column shows the probability maps of the model using dense (full) labels for training and the last column shows the ground truth images.}
\label{sf2}
\end{figure*}

We observe that the average per-class accuracy is worse for the Sift Flow dataset which is due to the fact that there are many classes with very few labeled samples in the training set which do not have the power to represent their corresponding class and get smoothed out by the Total Variation loss. For instance out of 33 classes in training set for the case with 10 labeled pixel per training image, 18 classes out of 33 have less than 100 supervised labeled pixels for training which is not sufficient, and 4 of these 18 classes having zero training samples. This unbalance in the sparse training data results in Total Variation loss taking over and reducing the average per-­class accuracy. To show the utility of our algorithm we calculate average per-­class accuracy for the top 10 classes which form $92.77 \%$ of training data and have a more uniform distribution when sparsely sampled in Table (\ref{sf_3}).

\setlength{\tabcolsep}{4pt}
\begin{table}[H]
\begin{center}
\caption{Average per-class accuracy for the top 10 classes of Sift Flow dataset. The reported numbers are average of 5 experiments }
\label{sf_3}
\begin{tabular}{ccc}
\hline\noalign{\smallskip}
\# of labeled pixels  & supervised & semi-supervised \\
 per image & approach & approach \\
\noalign{\smallskip}
\hline
\noalign{\smallskip}
10  & $ 48.53  $ & $ 58.19  $ \\
20  & $ 51.38  $ & $ 60.20  $ \\
30  & $ 54.25  $ & $ 60.38  $ \\
40  & $ 55.71  $ & $ 59.92 $ \\
50  & $ 56.80  $ & $ 59.85  $ \\
\hline
\end{tabular}
\end{center}
\end{table}
\setlength{\tabcolsep}{1.4pt}

\section{Conclusion}

In this paper we proposed to enforce spatial constraints on the output probability maps of learning algorithms. These spatial constraints help the learning system in converging to a better minimum in the solution space, especially in the case when we do not have enough labeled data. We proposed and showed how to minimize these constraint through learning via backpropagation in any pixel classifier. Our approach can be useful for learning deep structures like ConvNets, since deep structures need large amount of training data for a good performance. The proposed method uses  the sparse labeled data to avoid any trivial solution to the unsupervised loss function. From the probabilistic point of view we can interpret the spatial constraint proposed in this paper as a prior probability on the learning system parameters that restrict the solution to some smooth functions. Considering a graphical model paradigm, we can interpret the spatial constraints as the pairwise energy term in classical CRF models, where the unary energy term represents the supervised energy.\\

We showed significant improvement on three different datasets and two different pixel classifiers when only sparse labels are available. It is noteworthy that focus of this paper is to show practicality and effectiveness of the spatial constraints for sparse labels, not proposing an algorithm to achieve state of the art pixel accuracy on the task of semantic segmentation. We should point out the results reported in this paper can be further improved by selecting better models or selecting sparse labels by an expert who can help the learning algorithm by selecting informative pixels to label rather than randomly labeling pixels, as we have done in the paper.\\

Finally, we note that our framework is general and can be used to construct other unsupervised loss functions based on spatial characteristics of different images. For instance, substantial work has been carried out to enhance curvilinear and tubular structures in classical image processing. These methods typically make use of the first-order local derivative structure, i.e. Structure Tensor (\cite{weickert1999coherence}), or second-order local derivative structure, i,e. Hessian (\cite{frangi1998multiscale,tasdizen2005enhancement}) of images containing curvilinear and tubular structures. Both of these approaches can be formulated as unsupervised loss functions for learning following the general methodology introduced in Section~\ref{sec:unsupgeneral}.

\section{Acknowledgements}
This work was supported by NSF IIS-1149299.


\section{References}

  \bibliographystyle{elsarticle-harv} 
  \bibliography{egbib.bib}




%%\bibitem[ ()]{}

%%\end{thebibliography}
\end{document}